\pdfoutput=1
\documentclass[11pt]{article}

\usepackage[preprint]{acl}

\usepackage{times}
\usepackage{latexsym}

\usepackage[T1]{fontenc}

\usepackage[utf8]{inputenc}

\usepackage{microtype}

\usepackage{inconsolata}

\usepackage{listings}

\usepackage{graphicx}

\usepackage{todonotes}

\usepackage{enumitem}

\usepackage{booktabs}
\usepackage{multirow}
\usepackage[table]{xcolor}
\usepackage{tabularx}

\usepackage[ruled,vlined]{algorithm2e}

\usepackage{comment}

\usepackage{amsmath}
\usepackage{amssymb}  

\lstdefinestyle{prompt}{
  basicstyle=\ttfamily\footnotesize,
  columns=fullflexible,
  frame=single,
  breaklines=true,
  breakatwhitespace=true,
  showstringspaces=false,
  xleftmargin=1.5em,
  framexleftmargin=1.2em
}
\newcommand{\HierVA}{\textsc{HierVA}}
\newcommand{\HierVAFull}{Hierarchical Visual Agent (\HierVA)}

\title{Hierarchical Visual Agent: Managing Contexts in Joint Image-Text Space for Advanced Chart Reasoning}

\author{
  \textbf{Qihua Dong\textsuperscript{1}},
  \textbf{Ruozhen He\textsuperscript{2}},
  \textbf{Junwen Chen\textsuperscript{3}},
  \textbf{Yizhou Wang\textsuperscript{1}},
  \textbf{Xu Ma\textsuperscript{3}},
  \textbf{Songyao Jiang\textsuperscript{3}},
  \textbf{Yun Fu\textsuperscript{1}}
\\
\\
  \textsuperscript{1}Northeastern University,
  \textsuperscript{2}Rice University,
  \textsuperscript{3}Amazon AGI
}

\begin{document}
\maketitle

\begin{abstract}
Advanced chart question answering requires both precise perception of small visual elements and multi-step reasoning across several subplots.
While existing MLLMs are strong at understanding single plots, they often struggle with multi-step reasoning across multiple subplots.
We propose \HierVA{}, a \textbf{hierarchical visual agent} framework for chart reasoning that iteratively constructs and updates a working context in a joint image--text space. A high-level manager generates plans and maintains a compact context containing only key information, while specialized workers perform reasoning, gather evidence, and return results.
In particular, the agent maintains separate visual and textual contexts, using a \emph{zoom-in tool} to restrict the visual context.
Experiments on the \textsc{CharXiv} reasoning subset demonstrate consistent improvements over strong multimodal baselines, and ablation studies verify that hierarchical architecture, scoped visual context, and distilled context contribute complementary gains.
\end{abstract}

\section{Introduction}
\label{sec:intro}

\begin{figure}[t]
    \centering
    \includegraphics[width=\columnwidth]{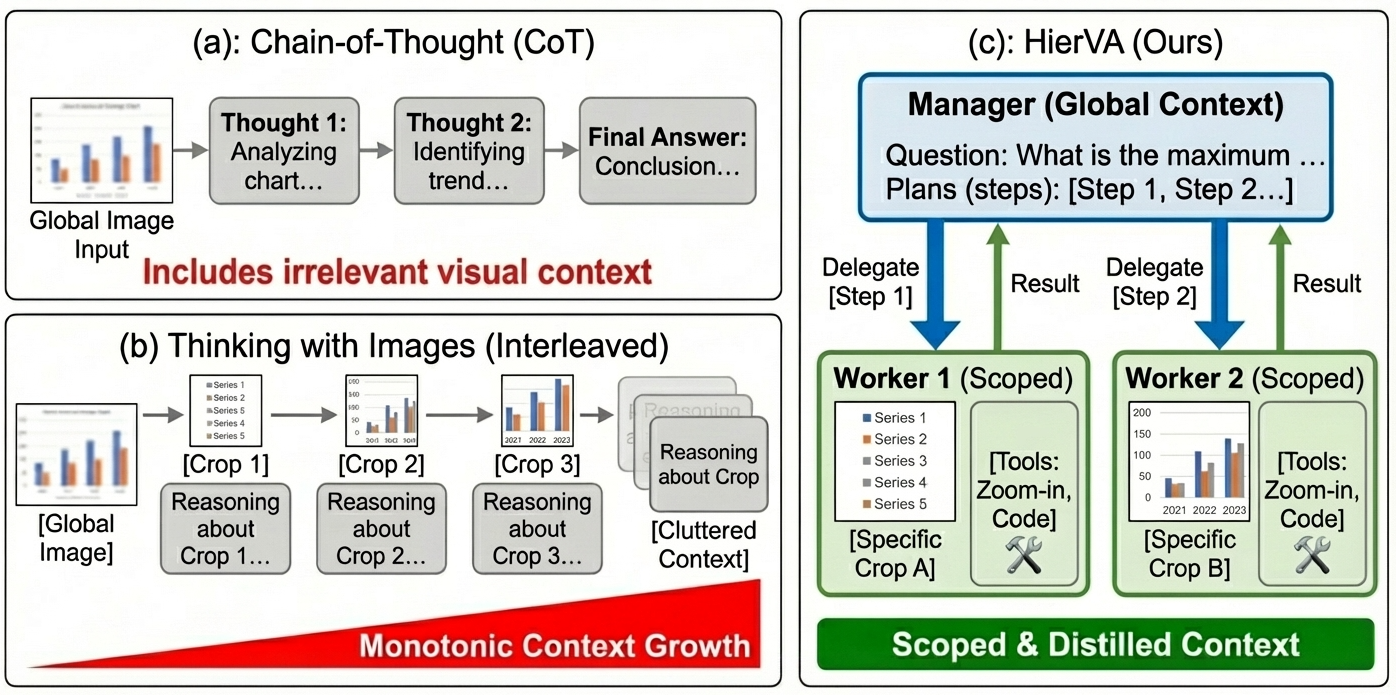}
    \caption{\textbf{Comparison of chart reasoning paradigms.} (a) \textbf{Chain-of-Thought (CoT)} reasons sequentially over a global image but is often distracted by irrelevant visual content.
    (b) \textbf{Thinking with Images (Interleaved)} iteratively acquires visual crops but appends them to a monotonically growing context, leading to clutter and distraction.
    (c) \textbf{\HierVA{} (Ours)} uses hierarchical visual agents that maintain a joint image-text working context. Each subtask operates in a scoped local context with task-specific crops and tasks, returning only distilled results to prevent context overload.}
    \label{fig:teaser}
\end{figure}

\begin{table*}[t]
\centering
\small
\begin{tabular}{@{}p{0.08\textwidth} p{0.42\textwidth} p{0.42\textwidth}@{}}
\toprule
& \textbf{Basic Chart Understanding} & \textbf{Advanced Chart Reasoning} \\
\midrule
\textbf{Image} &
\raisebox{-0.5\height}{\includegraphics[width=0.4\textwidth]{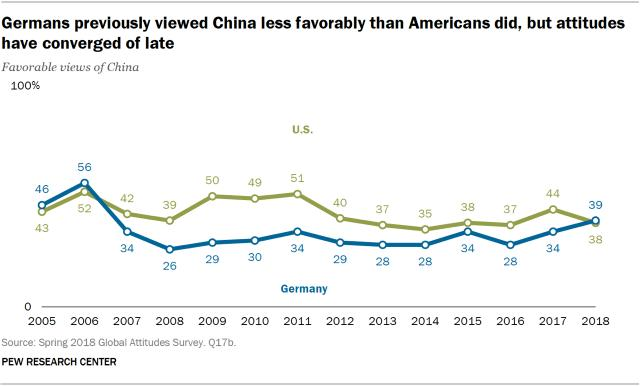}} &
\raisebox{-0.5\height}{\includegraphics[width=0.4\textwidth]{figures/basic_vs_advanced/advanced.png}} \\
\midrule
\textbf{Question} &
What's the color of graph with 56 as the highest value? &
In the graphs, how many times does the log10-Distance value for TMR fall within the range of [-2.5, -2.0]? \\
\bottomrule
\end{tabular}
\caption{Comparison of basic vs.\ advanced chart reasoning tasks. Basic tasks require localizing a single region and one-step retrieval. Advanced tasks require aggregating evidence across multiple regions and multi-step computation.}
\label{tab:basic_vs_advanced}
\end{table*}

\emph{Chart question answering} is a key capability for scientific assistants, document understanding systems, and accessibility tools.
While existing multimodal large language models (MLLMs) are strong at understanding single plots, they often struggle with multi-step reasoning across multiple subplots~\citep{wang2024charxiv}.
This hinders their application in real-world chart reasoning tasks.

The core challenge is that complex chart reasoning is inherently an \emph{image-text mixed task}: answering a single question often requires aggregating fine-grained visual details---tick labels, axis units, legend entries, dense markers---from multiple scattered regions.
This creates two compounding difficulties.
First, relevant evidence is often small or cluttered when the chart is viewed at global scale, making accurate recognition unreliable.
Second, multi-step reasoning (e.g., comparing series across coordinates, computing differences, or identifying constrained extrema) demands retaining and coordinating intermediate results across steps.
As the reasoning chain grows, the accumulated context becomes long and heterogeneous, mixing task-relevant signals with information that is irrelevant to the current step---causing earlier evidence to be ``forgotten'' or diluted.

Multiple approaches have been proposed to manage context in chart reasoning tasks.
For multi-step reasoning, text-based chain-of-thought (CoT) prompting~\citep{wei2022chainofthought,yao2022react,wang2023plan,shinn2023reflexion,zhang2025recursive} is mainly adopted to elicit explicit reasoning traces.
Recent ``thinking with images'' patterns incorporate visual details into reasoning traces by acquiring additional visual evidence (e.g., zooming into a region) during inference~\citep{openai2025o3systemcard,lai2025minio3,zheng2025deepeyes}, making it easier for the agent to focus on the current step's region of interest.
However, a common failure mode is \emph{monotonic context growth}: since images consume tokens, repeatedly appending crops and intermediate text dilutes the global context and degrades later steps that rely on it---especially those requiring access to the original visual context.

This paper advocates a simple but consequential principle: \emph{multimodal reasoning benefits from managing contexts in joint image-text space with the same discipline used to manage textual reasoning traces}.
Concretely, effective chart agents should
(i) treat visual context with the same care as textual context,
(ii) scope each step's context to only the information necessary for that step, and
(iii) distill intermediate traces into compact, high-signal state.
In other words, instead of continuously growing the context, the system should \emph{refine and compress} it, keeping the conditioning information focused and stable across steps.

We operationalize this principle with \textbf{\HierVAFull}, a training-free manager--worker framework for advanced chart reasoning.
The manager performs \textbf{planning and zoom-in} to restrict textual and visual contexts to each step's needs, dispatching them as explicit inputs to the appropriate worker.
Crucially, our hierarchy enforces \textbf{context distillation and encapsulation}: intermediate artifacts (e.g., multiple planning drafts) are refined and only the clean final version is retained in the manager context; each worker's reasoning trace is hidden behind an abstraction barrier so the manager observes only the assigned task and its result; and workers operate with minimal, task-specific contexts, improving efficiency and reducing distraction.
This design keeps the global context compact and high-signal while preserving rich intermediate computation locally where it is needed.

We evaluate on the \textsc{CharXiv} reasoning subset and show consistent improvements over strong multimodal baselines.
To summarize, our contributions are:
\begin{itemize}[noitemsep]
    \item A \textbf{manager-worker hierarchical visual agent} (\HierVA) with explicit multimodal working context.
    \item \textbf{Context distillation and encapsulation} to improve efficiency and reduce distraction.
    \item \textbf{Improved performance} on advanced chart reasoning tasks without additional training.
\end{itemize}

\section{Method}
\label{sec:method}

\subsection{Problem setup}
Given a chart image $I_0$ and a natural-language question $q$, the goal of chart question answering (QA) is to produce an answer $a$.

\begin{algorithm}[t]
\caption{\HierVA{} for chart task}
\label{alg:hva}
\KwIn{Chart image $I_0$, question $q$, skill library $\mathcal{S}$, tools $\mathcal{T}$}
\KwOut{Answer $a$}
Initialize manager context $C_M \leftarrow \{q, I_0\}$\;
$P \leftarrow \textnormal{\textsc{Plan}}(C_M)$ with optional refinement; store only refined $P$ in $C_M$\;
\For{$t = 1,2,\dots$}{
  \If{$t>1$ \textnormal{and} $\textnormal{\textsc{ShouldStop}}(C_M)$}{
    \Return $\textnormal{\textsc{FinalAnswer}}(C_M)$\;
  }
  $\tau_t \leftarrow \textnormal{\textsc{CreateNextTask}}(C_M)$ using schema-constrained self-correction\;
  $o_t \leftarrow \textnormal{\textsc{ExecuteWorker}}(\tau_t)$ with one image and selected skills/tools\;
  \If{$\textnormal{\textsc{HasHook}}(\tau_t)$}{
    $o_t \leftarrow \textnormal{\textsc{ExecuteHook}}(o_t)$\;
  }
  Append a distilled summary of $(\tau_t, o_t)$ to $C_M$\;
}
\end{algorithm}

\subsection{\HierVA: Hierarchical Visual Agent}
\label{sec:hva}
We implement the principle from \S\ref{sec:intro}---\emph{acquire, scope, and distill evidence}---with \textbf{\HierVA}, composed of a \textbf{manager} and multiple \textbf{workers}. Conceptually, both manager and workers are instances of the same underlying multimodal model, but they are used with different \emph{roles}, \emph{tools}, and \emph{context}.

\paragraph{Global vs.\ local contexts.}
The manager maintains a persistent global working context $C_M$ (a message history) containing:
(i) the original question and chart reference, (ii) a refined plan, and (iii) compact, structured summaries of completed subtasks.
In contrast, each worker is invoked with an isolated local context $C_{W_t}$ containing only:
(i) the task instruction and minimal required background, (ii) an optional \emph{skill} procedure selected by the manager (see \S\ref{sec:skill_routing}), and (iii) a \emph{single} image input (either $I_0$ or a zoomed crop).
This separation enforces \textbf{encapsulation}: the manager never appends the worker’s deliberation trace into $C_M$, preventing irrelevant intermediate text from accumulating across steps.

\paragraph{Action space.}
At each step $t$, the manager chooses between (a) requesting additional \emph{visual evidence} via a zoom-in crop, or (b) requesting a \emph{textual fact} (e.g., a read-off value, a comparison result, an intermediate computation). Zoom actions are executed by workers through a dedicated image tool, and the cropped image will be returned to the manager; textual subtasks may additionally use lightweight tools (e.g., code execution) chosen by the manager.

\subsection{Manager control loop}
\label{sec:manager_loop}
Algorithm~\ref{alg:hva} summarizes the orchestration loop. The process begins with a two-stage planning phase, then iterates over task creation, worker execution, optional coordinate normalization, and distilled reporting back to the manager until termination.

\paragraph{Two-stage planning with refinement.}
The manager first prompts for a high-level plan, then prompts for an explicit step list.
The plan is critical for the final results since it lists clearly the context for each worker and we find the two-stage planning is more effective.
If we ask for a detailed explicit plan at once, the manager may struggle to separate context for workers in a clean way.
Crucially, the manager retains only the \emph{final refined plan} in $C_M$, discarding intermediate drafts.

\paragraph{Iterative execution and termination.}
After the first task is executed, the manager periodically checks whether it has enough information to answer the original question. If not, it generates the next subtask, dispatches an appropriate worker, and incorporates the resulting distilled fact into $C_M$. When the manager predicts that no further subtasks are needed, it produces the final answer in a standardized format (we use a \texttt{\textbackslash boxed\{\}} wrapper for reliable extraction).

\subsection{Adaptive zoom and worker execution}
\label{sec:zoom}
Fine-grained chart evidence is often not readable from a single global-scale encoding of $I_0$. We therefore treat zooming as an explicit action: the manager can create an \texttt{image}-expected task whose worker calls a zoom-in tool. A zoom action specifies a bounding box and gets a cropped, resized image for high-resolution inspection.

Each worker is created on demand with exactly one image input (either the original chart or a previously generated crop), enforcing scoped evidence. Workers produce outputs matching the expected type; for text tasks, we extract only the final sentence to keep manager-visible updates compact (see Appendix~\ref{sec:impl_details} for details on task schema and image indexing).

\subsection{Skill routing and context management}
\label{sec:skill_routing}

For certain chart images, code execution can yield more accurate answers (e.g., extracting precise point coordinates). However, workers tend to respond directly rather than invoking code tools. To address this, we introduce task-specific \emph{skills} into the worker's context.
Naively encoding all such procedural knowledge in the manager or worker's base prompt would bloat $C_M$ and encourage monotonic context growth. Instead, we maintain a compact \textbf{skill library} $\mathcal{S}$, where each skill is a short markdown procedure. For each task, the manager selects a relevant subset of skills and injects their text into the worker's system prompt.
Crucially, the manager never reads the skill content, and workers only see skills when explicitly selected.
This \textbf{just-in-time} injection guides workers effectively while avoiding context bloat---a direct benefit of the \HierVA{} design.

Our orchestration explicitly targets monotonic context growth through three mechanisms:
\textbf{(1) Plan distillation}: only the final refined plan is retained, discarding intermediate drafts.
\textbf{(2) Worker encapsulation}: workers execute in private contexts; the manager does not inherit worker message histories or tool call traces.
\textbf{(3) Result distillation}: worker responses are converted to compact summaries (a single sentence for text tasks, a crop handle for image tasks) before being added to $C_M$.
Together, these mechanisms preserve rich intermediate computation locally while keeping global conditioning focused and stable.

\section{Basic Chart Understanding Tasks}
\label{sec:basic_chart_tasks}

In this section, we examine why text-based CoT and ``thinking with images'' alone are insufficient for chart tasks, and argue that explicit visual context management is a critical component of chart reasoning.

\subsection{Initial Study on ChartQA}
On classic benchmarks such as ChartQA~\citep{masry2022chartqa}, current MLLMs with text-based CoT already achieve near-saturated performance (90\%+~\citealp{bai2025qwen3vl}), while ``thinking with images'' approaches~\citep{zheng2025deepeyes} offer little additional gain.
This is unsurprising: ChartQA images are relatively simple, typically containing a single plot with few visual elements, so the visual context remains short and requires no explicit management.

\subsection{Task Setup}
We consider two chart understanding settings: (i) the \textsc{ChartQA} validation set (1.92k examples), where each example typically consists of a single plot requiring single-step reasoning (similar to the left sample in Figure~\ref{tab:basic_vs_advanced}), and (ii) synthetic composite charts (1k examples), where we systematically control chart complexity by varying the number of subplots (sp\#) and data series (see \S\ref{par:synthetic_data_collection} for generation details).
We report accuracy (\%) and use the same backbone model across all methods.

\begin{table}[t]
\centering
\small
\resizebox{\columnwidth}{!}{%
\begin{tabular}{@{}l|c|cccc@{}}
\toprule
\multirow{2}{*}{\textbf{Method}} & \textbf{ChartQA} & \multicolumn{4}{c}{\textbf{Synthetic Charts}} \\
 & sp\#1 & sp\#1 & sp\#2 & sp\#4 & sp\#6 \\
\midrule
Direct & 88.9 & 88.5 & 86.5 & 84.8 & 83.1 \\
CoT & \textbf{90.2} & \textbf{90.1} & \textbf{89.2} & 88.4 & 87.5 \\
CoT-Plan & 90.1 & 89.8 & 88.1 & 86.2 & 86.0 \\
\midrule
Thinking w/ Images & 89.9 & 89.7 & 88.5 & 87.4 & 87.5 \\
\rowcolor{blue!8}
\HierVA~(Ours) & 89.9 & 89.7 & \textbf{89.2} & \textbf{88.5} & \textbf{88.2} \\
\bottomrule
\end{tabular}%
}
\caption{Comparison of methods on ChartQA and synthetic charts with varying number of subplots (sp\#), using Qwen3VL-A22B. The metric is accuracy(\%).}
\label{tab:chartqa_comparison}
\end{table}

\subsubsection{Synthetic Charts}
In order to fairly study the performance of different methods on chart tasks when the input images are of different complexity, we use GPT5~\citep{openai2025gpt5systemcard} to synthesize chart data, and control the number of subplots and the number of visual elements in the input image.

\paragraph{Data Collection}
\label{par:synthetic_data_collection}
We use few-shot prompting to generate questions that remain ChartQA-like in style and difficulty.
For input images, we cover diverse chart types (line, bar, scatter, pie, etc.), varying numbers of data series (2--10), and different subplot counts (1, 2, 4, 6).
For each question--plot pair, we generate four input images with 1, 2, 4, and 6 subplots respectively; the target plot is randomly placed in one subplot while the remaining subplots are filled with randomly generated charts. This controlled setup enables systematic study of the impact of visual context complexity.

\subsection{Methods}
We compare against four baselines using the same backbone model Qwen3VL-A22B~\citep{bai2025qwen3vl}.

\paragraph{Baselines}
\label{par:baselines}

\textbf{Direct} prompts the model with the question and chart image without any
chain-of-thought reasoning. \textbf{CoT} follows standard chain-of-thought
prompting~\citep{wei2022chainofthought}, instructing the model to reason
step-by-step before producing the final answer. \textbf{CoT-Plan}~\citep{wang2023plan} uses a ``plan then solve'' prompt, first generating a multi-step plan decomposing the question, then executing each step sequentially to derive the final answer. Exact prompts for these baselines are provided in Appendix~\ref{sec:prompting}. \textbf{Thinking w/ Images}
implements the recently proposed ``thinking with images''
paradigm~\citep{zheng2025deepeyes,lai2025minio3}, where the model can
iteratively acquire additional visual evidence (zoom-in crops) during
reasoning. Unlike \HierVA, this baseline appends all intermediate crops and
reasoning traces to a single growing context without hierarchical delegation
or context distillation.

\paragraph{\HierVA}
\HierVA~is our proposed method described in \S\ref{sec:method}. For this study,
we use the same backbone model and provide the zoom-in action. No other tools are used, and there are no injected skills.
The key differences from ``Thinking w/ Images'' are: (i) hierarchical delegation where
a manager dispatches subtasks to workers with isolated contexts, (ii) context
distillation that retains only compact summaries rather than full reasoning
traces, and (iii) scoped evidence where each worker receives only the relevant
image crop rather than accumulating all crops in a single context.

\paragraph{Action Space} For \HierVA~and \textbf{Thinking w/ Images}, we provide a `zoom-in' action.

\subsection{Results and Analysis}
Table~\ref{tab:chartqa_comparison} summarizes the performance of all methods on ChartQA and synthetic charts with varying subplot counts.

\paragraph{Composite charts degrade accuracy.}
Across all methods, accuracy consistently drops as the number of subplots
increases: from sp\#1 to sp\#6, we observe a decline of 5.4\% for Direct,
2.6\% for CoT, 3.8\% for CoT-Plan, 2.2\% for Thinking w/ Images, and 1.5\% for \HierVA.
This trend confirms that composite chart images pose a fundamental challenge:
the target information occupies a smaller fraction of the visual input,
and the other visual context makes it harder for models to answer the question.

\paragraph{Managing visual context helps.}
Methods with zoom capability (\textbf{Thinking w/ Images} and \HierVA)
degrade more gracefully than \textbf{Direct}, \textbf{CoT}, and \textbf{CoT-Plan}, showing
that actively managing visual context---by acquiring high-resolution crops
of relevant regions---helps preserve accuracy on composite charts.

\paragraph{Context management does not yet differentiate.}
Interestingly, \HierVA~and \textbf{Thinking w/ Images} perform similarly
on these ChartQA-style questions despite their architectural differences.
We attribute this to: most questions here require
single-step value retrieval rather than multi-step reasoning, so the
benefits of context management do not yet manifest.

\paragraph{Motivation for complex reasoning tasks.}
These results suggest that for simple retrieval questions, \emph{acquiring}
evidence suffices---but we hypothesize that \emph{managing} textual and visual context becomes
critical when reasoning chains grow longer and intermediate results must be
tracked. This motivates our evaluation on \textsc{CharXiv} (\S\ref{sec:charxiv}),
where questions require multi-step comparisons, aggregations, and computations.

\section{Advanced Chart Reasoning Tasks}
\label{sec:charxiv}

Real-world chart tasks are often far more complex than the basic retrieval questions studied in \S\ref{sec:basic_chart_tasks}.
To evaluate advanced, multi-step chart reasoning in the wild, we turn to the \textsc{CharXiv} reasoning split~\citep{wang2024charxiv}, a challenging benchmark that requires multi-step comparisons, aggregations, and reasoning over composite charts drawn from arXiv papers.
Even strong closed-source models struggle on this benchmark, making it an ideal testbed for our approach~\citep{wang2024charxiv}.

\paragraph{Difference between basic and advanced chart tasks.}
An example is shown in Table~\ref{tab:basic_vs_advanced}.
In basic chart tasks (\S\ref{sec:basic_chart_tasks}), most questions can be answered by performing one-step value retrieval.
In contrast, advanced chart questions often require (i) aggregating evidence across multiple regions/series or panels, and (ii) maintaining intermediate results to complete multi-step computations (e.g., deltas, ratios, averages).
This shift makes long-horizon \emph{context management} critical: naive approaches that append all intermediate reasoning and visual evidence into one growing transcript are more likely to lose key facts or introduce inconsistencies as the chain length increases.

\begin{table*}[t]
\centering
\small
\setlength{\tabcolsep}{3pt}
\begin{tabularx}{0.9\textwidth}{Xcc|c|cccccc|c}
\toprule
\multirow{2}{*}{\textbf{Method}} & \multirow{2}{*}{\textbf{Image Tools}} & \multirow{2}{*}{\textbf{Skills}} & \multicolumn{7}{c|}{\textsc{CharXiv}} & \multirow{2}{*}{\textbf{Peak Tok \#~}} \\
\cmidrule(lr){4-10}
& & & \textbf{All} & \textbf{Extr} & \textbf{First} & \textbf{Read} & \textbf{RevR} & \textbf{Comp} & \textbf{Freq} & \\
\midrule
Direct & --- & --- & 45.7 & 38.5 & 42.1 & 75.4 & 62.3 & 28.9 & 27.0 & 702.48 \\
CoT & --- & --- & 62.1 & 64.2 & 53.5 & \textbf{78.9} & 69.1 & 54.4 & 52.5 & 1926.19 \\
CoT-Plan & --- & --- & 62.4 & 64.8 & 54.1 & 78.2 & 69.4 & 55.1 & 52.8 & 1947.21 \\
\midrule
Thinking w/ Images & zoom & --- & 58.9 & 63.5 & 52.1 & 77.5 & \textbf{70.2} & 45.1 & 43.0 & 2995.02 \\
Thinking w/ Images & zoom, code & --- & 50.1 & 48.5 & 44.2 & 72.5 & 63.1 & 38.5 & 33.8 & 2990.23 \\
Thinking w/ Images & zoom, code & fixed & 55.3 & 58.2 & 48.5 & 76.8 & 66.5 & 42.1 & 39.7 & 2837.14 \\
\midrule
\rowcolor{blue!8}
\textbf{HierVA (Ours)} & zoom, code & dynamic & \textbf{64.2} & \textbf{66.3} & \textbf{54.5} & 78.4 & {70.1} & \textbf{59.9} & \textbf{55.0} & {2568.38} \\
\bottomrule
\end{tabularx}
\caption{Performance on \textsc{CharXiv} reasoning subset. Accuracy (\%) by question type: Extr=Find extreme, First=Find the first, Read=Read value, RevR=Reverse read, Comp=Compare, Freq=Frequency count. Peak Tok \#=avg.\ per-example peak token count.}
\label{tab:charxiv_comparison}
\end{table*}

\begin{table}[t]
\centering
\resizebox{\columnwidth}{!}{%
\begin{tabular}{@{}lcccc|c@{}}
\toprule
\textbf{Variant} & \textbf{Hier.} & \textbf{Dist.} & \textbf{Scop.} & \textbf{Skill} & \textbf{Acc.\ ($\downarrow$)} \\
\midrule
\rowcolor{blue!8}
\textbf{Full \HierVA} & \checkmark & \checkmark & \checkmark & \checkmark & \textbf{64.2} \\
\quad w/o hierarchy & $\times$ & \checkmark & \checkmark & \checkmark & 53.8 {\small($\downarrow$10.4)} \\
\quad w/o distillation & \checkmark & $\times$ & \checkmark & \checkmark & 55.5 {\small($\downarrow$8.7)} \\
\quad w/o scoped context & \checkmark & \checkmark & $\times$ & \checkmark & 61.1 {\small($\downarrow$3.1)} \\
\quad w/o dynamic skills & \checkmark & \checkmark & \checkmark & $\times$ & 59.6 {\small($\downarrow$4.6)} \\
\bottomrule
\end{tabular}%
}
\caption{Ablation study on \textsc{CharXiv} reasoning subset. Hier.=hierarchical delegation, Dist.=context distillation, Scop.=scoped context, Skill=just-in-time skill routing.}
\label{tab:charxiv_ablation}
\end{table}

\subsection{Task Setup}
\paragraph{\textsc{CharXiv} reasoning.}
We evaluate on the validation split of the \textsc{CharXiv} reasoning subset, which contains 1000 examples~\citep{wang2024charxiv}, and report \textbf{accuracy}.
For finer-grained analysis, we use an LLM to classify questions into six reasoning types: \emph{read value}, reading or estimating a numeric value; \emph{find extreme}, identifying max/min or ranked extremes; \emph{find the first}, identifying the first crossing, drop, plateau, or threshold event; \emph{reverse read}, selecting the label, method, or subplot matching a described pattern; \emph{compare}, comparing two items; and \emph{freq}, counting occurrences or items meeting a condition.

\subsection{Methods}

\paragraph{Baselines.}
For \textbf{Direct}, \textbf{CoT}, and \textbf{CoT-Plan}, we use the same tool-free baselines as in \S\ref{par:baselines}.

To fairly isolate the effects of tool access and skill injection, we evaluate three variants of \textbf{Thinking w/ Images}:
(i) with the \emph{zoom-in tool only}, matching the baseline in \S\ref{par:baselines};
(ii) with \emph{zoom-in and code interpreter}, matching the tool set available to \HierVA; and
(iii) with \emph{zoom-in, code interpreter, and a fixed skill list} prepended to the system prompt.
In variant (iii), the skills mirror those used by \HierVA~(\S\ref{sec:skill_routing}), but are provided statically rather than injected dynamically per subtask.

\paragraph{\HierVA.}
Here, in addition to the zoom-in tool, \HierVA~can use a code interpreter to programmatically analyze zoomed chart crops (e.g., extracting series/marks by color, locating points/bars, reading values, comparing values, and identifying extrema), and to carry out the corresponding derived computations needed for multi-step questions. \HierVA~also uses skills as compact procedural guidance: for each subtask, the manager selects a task-relevant subset from the skill library and injects it into the worker's system prompt just in time (see \S\ref{sec:skill_routing}), improving reliability without bloating the global manager context.

Concretely, we use five injected skills (shown as short markdown documents) including:
\texttt{ReadValue} (extract and parse a numeric value),
\texttt{FindExtremum} (argmax/argmin over extracted values),
\texttt{CompareValues} (pairwise comparison with unit/format normalization),
\texttt{ComputeAggregate} (sum/mean/range over extracted values),
and \texttt{ComputeRatioDelta} (ratios/percent change/differences).

\subsection{Case Study}
\begin{figure*}[t]
    \centering
    \includegraphics[width=\textwidth]{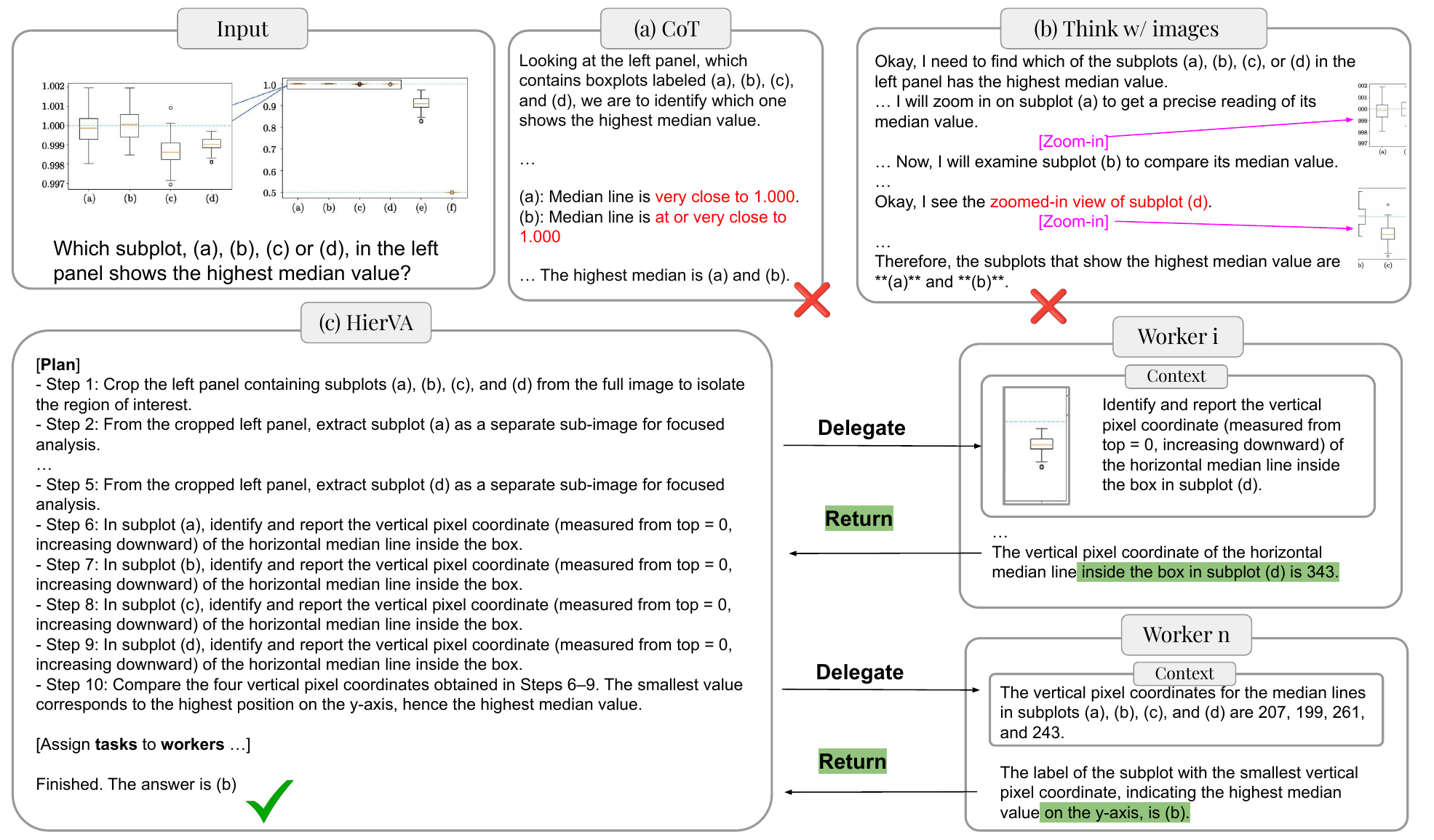}
    \caption{\textbf{Case study comparing reasoning approaches.} We illustrate how different methods handle an advanced chart reasoning question. \textbf{CoT} reasons sequentially over the global image but struggles with fine-grained details. \textbf{Thinking w/ Images} iteratively zooms into regions but accumulates all crops and intermediate reasoning in a growing context, leading to errors. \textbf{\HierVA{}} manages the context effectively by using hierarchical agents and scoped context.
    Red text highlights incorrect reasoning steps.}
    \label{fig:case_study}
\end{figure*}

Figure~\ref{fig:case_study} presents an example where the model must answer an advanced chart reasoning question: among four closely spaced boxplots (a--d), identify the sub-plot with the highest median. \textbf{CoT} attempts to judge medians directly from the global view and incorrectly concludes that (a) and (b) are tied, reflecting brittle estimation when differences are subtle.
\textbf{Thinking w/ Images} improves perception by zooming, but because it appends every crop and intermediate rationale into a single growing context, it becomes prone to bookkeeping errors---for instance, when attempting to crop subplot (d), it mistakenly crops subplot (c)---and ultimately produces an incorrect answer.
In contrast, \textbf{\HierVA{}} decomposes the task into scoped worker subtasks: the manager plans the necessary steps, including cropping the left panel and each sub-plot, and delegates each sub-plot to a dedicated worker with a single relevant view. Each worker localizes the median line and returns a compact scalar measurement (the median line's vertical pixel coordinate).
The manager then compares these returned measurements and correctly identifies (b) as the highest-median sub-plot.

\subsection{Results and Analysis}

\paragraph{Main results.}
Table~\ref{tab:charxiv_comparison} presents the results on the \textsc{CharXiv} reasoning subset. \HierVA{} achieves the highest overall accuracy of \textbf{64.2\%}, surpassing the strongest baseline (CoT-Plan) by 1.8\% and the Thinking w/ Images baseline by 5.3\%. Notably, our method shows the largest gains on complex reasoning categories such as \textit{Find Extreme} (+1.5\%), \textit{Compare} (+4.8\%), and \textit{Frequency} (+2.2\%), where multi-step evidence management is most critical.

We also compare against tool-augmented variants of Thinking w/ Images. Interestingly, naively enabling code execution degrades performance (50.1\%): without procedural guidance, the model often generates incorrect or irrelevant code. Providing a fixed skill list in the system prompt partially recovers accuracy (55.3\%), but still falls short of \HierVA{}. This confirms that \emph{how} tools and skills are managed---via dynamic routing and context distillation---is as important as having access to them.

\paragraph{Efficiency and token usage.}
We compare the \textit{average peak token count}---the maximum number of tokens in the context window during an inference episode, averaged across all examples.
As shown in the rightmost column of Table~\ref{tab:charxiv_comparison}, \textbf{Thinking w/ Images} suffers from rapid context growth (approx. 3000 tokens) as it blindly appends visual and textual history. In contrast, \HierVA{} maintains a more compact working context (2568 tokens, $\sim$14\% reduction) despite executing complex tool-use chains. This reduction confirms that our context distillation and encapsulation mechanisms effectively prune irrelevant context.

\paragraph{Ablations.}
We ablate the key components of \HierVA to quantify their contributions on advanced reasoning (Table~\ref{tab:charxiv_ablation}).
Specifically, we ablate four elements: \textbf{Hierarchical delegation} (manager--worker decomposition; \S\ref{sec:hva}), \textbf{context distillation} (keeping only refined plans and compact summaries; \S\ref{sec:manager_loop}), \textbf{scoped context} (each worker sees a single relevant image view; \S\ref{sec:zoom}), and \textbf{just-in-time skill routing} (injecting only task-relevant procedural skills; \S\ref{sec:skill_routing}).
The results show that \textbf{Hierarchical Delegation} is the most critical factor, with a 10.4\% drop when removed, confirming that decomposing long-horizon tasks into manager-worker interactions is essential.
\textbf{Context Distillation} also proved vital (8.7\% drop), as appending full reasoning traces without summarization severely degrades performance on multi-step chains.
\textbf{Just-in-time Skill Routing} (4.6\% drop) and \textbf{Scoped Context} (3.1\% drop) provide further complementary gains by keeping procedural guidance relevant and visual evidence focused.

\section{Related Work}
\label{sec:related}

\paragraph{Chart understanding and chart reasoning.}
Chart QA couples fine-grained perception (ticks, legends, dense marks) with multi-step numeric reasoning. Recent work improves chart QA by converting plots into structured text for downstream language-model reasoning \citep{deplot2023}, or by training chart-specialized vision--language models via derendering and chart-aware pretraining \citep{matcha2023,unichart2023}. Large-scale chart instruction tuning further broadens coverage of real-world questions \citep{chartinstruct2024,chartassistant2024,chartgemma2024}, and reasoning-oriented objectives target harder multi-step inference \citep{chen2025chartr1}. Realistic evaluations still reveal substantial gaps for multimodal LLMs on scientific charts \citep{wang2024charxiv}, motivating agentic approaches that decompose queries into visually grounded subtasks \citep{kaur2025chartagent}. Our work complements these directions by focusing on inference-time orchestration—adaptive zoom, skill routing, and context distillation/encapsulation to control multimodal working context.

\paragraph{Thinking with Images: Tool-Augmented Multimodal Reasoning.}
Beyond charts, a broad line of work treats multimodal problem solving as an \emph{agentic} loop in which an LLM plans, calls external vision modules/tools, and integrates observations \citep{visualchatgpt2023,visprog2023,vipergpt2023,mmreact2023}. Recent ``thinking with images'' paradigms emphasize multi-turn visual search behaviors---e.g., iteratively zooming into relevant regions to obtain high-resolution evidence when a global view is insufficient \citep{zheng2025deepeyes,lai2025minio3,openai2025o3systemcard}. In parallel, stronger multimodal backbones increasingly support higher-resolution understanding and longer multimodal contexts, partially reducing (but not eliminating) the need for explicit perception tooling \citep{llavanext2024,bai2025qwen3vl}. A practical challenge across tool-augmented multimodal agents is that observations (crops, OCR strings, intermediate tool outputs) are often appended into a single ever-growing transcript, which can reduce signal-to-noise over long trajectories. Our hierarchical visual agent targets this bottleneck directly: workers operate with \emph{scoped} local contexts (including a single image view), while the manager only retains distilled, structured results---preventing monotonic growth of irrelevant intermediate text and redundant visual evidence.

\paragraph{Planning and context management for LLM agents.}
Our design also builds on progress in planning and context control for language agents. Chain-of-thought prompting improves multi-step reasoning by eliciting intermediate rationales \citep{wei2022chainofthought}, and ReAct-style frameworks interleave reasoning with actions to support tool use and environment interaction \citep{yao2022react}. Subsequent work explores explicit plan--execute decomposition \citep{wang2023plan}, iterative self-improvement via reflection \citep{shinn2023reflexion}, and search over candidate reasoning trajectories \citep{treeofthoughts2023}. Efficiency-oriented designs such as ReWOO decouple reasoning from observation interleaving to reduce prompt length and repeated computation \citep{rewoo2023}, which is closely aligned with our goal of limiting unnecessary context growth during multi-step tool use. A further line studies long-horizon memory and working-context management, including externalized memory hierarchies and hierarchical working-memory mechanisms \citep{memgpt2023,hiagent2024,zhang2025recursive}, as well as decomposition via multi-agent/role-based collaboration \citep{autogen2023,camel2023}. Prior methods are predominantly text-centric and treat observations as cheap to append. In contrast, chart reasoning requires controlling \emph{joint image--text} context under tight budgets. Our approach extends hierarchical planning and memory ideas to multimodal settings by (i) enforcing worker encapsulation, (ii) treating zoomed visual evidence as a scoped resource, and (iii) distilling intermediate artifacts into compact manager-visible state.

\section{Discussion: Beyond Charts}
\label{sec:beyond_charts}

We choose chart reasoning as a testbed because it exposes the context-management bottleneck sharply, but the underlying principle of \emph{scoped, distilled joint image--text context} is not chart-specific. The same manager--worker decomposition, zoom-as-action, and result distillation apply to any fine-grained visual reasoning setting where evidence is small, spatially scattered, and must be aggregated across steps: dense document VQA, multi-image reasoning over dashboards or slide decks, and scientific imagery (microscopy, medical, remote-sensing) all share this pathology. What would change across domains is mainly the concrete skill instances and tool hooks. The architectural ingredients, namely hierarchical delegation, plan/result distillation, and scoped per-worker context, are domain-agnostic, and we view validating them in these broader settings as a natural next step.

\section{Conclusion}

We presented \HierVAFull, a manager--worker framework for chart question answering that treats inference as \emph{multimodal context management}, combining adaptive zoom-in, just-in-time skill routing, and strict distillation/encapsulation to keep the working context compact and high-signal.
On \textsc{CharXiv}, \HierVA{} achieves 64.2\% accuracy, outperforming strong baselines by 1.8--5.3 points, with the largest gains on complex multi-step questions.
We expect the principle of \emph{scoped, distilled multimodal context} to extend beyond charts to other fine-grained visual reasoning domains where global views obscure critical details.

\section{Limitations}

Our approach inherits limitations from both the underlying multimodal model and the tool-augmented agent setting. If the backbone (or tool pipeline) cannot reliably perceive fine-grained chart evidence (ticks, legends, units, dense marks), hierarchical orchestration cannot recover missing facts, and zoom choices may amplify errors by focusing on the wrong region or omitting needed context. Decomposition can also be brittle for questions requiring global constraints or cross-panel alignment that do not cleanly factor into independent subtasks. Skill routing is limited by an incomplete library and imperfect selection, and scaling to broader skill sets with reliable tool use remains challenging. Finally, multi-step manager--worker execution increases latency and engineering complexity relative to single-pass prompting, and our evaluation (ChartQA, synthetic composites, \textsc{CharXiv}) does not cover all real-world chart styles, languages, or document layouts.

\section{Ethics Statement}

This work focuses on inference-time orchestration for chart question answering and does not involve collection of new data or human subjects research.
We evaluate on existing public benchmarks (\textsc{ChartQA}, \textsc{CharXiv}) that contain charts from publicly available sources.
Potential positive impacts include improved accessibility for visually impaired users and more reliable scientific document understanding.
We acknowledge that, like all capable multimodal systems, our method could in principle be misused for extracting information from proprietary charts; however, the techniques we propose do not introduce new capabilities beyond those of the underlying vision--language model.
Computational costs are moderate: our hierarchical design reduces peak context length relative to baselines with comparable tool access.

\section{Use of Large Language Models in Writing}
Large language models were used in a limited manner to assist with language polishing and improving clarity in the writing of this paper. All technical content, experimental design, analysis, and conclusions were conceived, verified, and finalized by the authors. The use of LLMs did not affect the scientific claims or results of the work.

\bibliography{custom}

\appendix

\section{Detailed Prompting}
\label{sec:prompting}

This appendix provides the exact prompt suffixes used for each baseline method.

\paragraph{Direct.}
We append the instruction: \texttt{"You cannot reason or think. You must ONLY answer directly in \textbackslash boxed\{\}."}

\paragraph{CoT (Chain-of-Thought).}
We append the instruction: \texttt{"Let's reason step by step."}

\paragraph{CoT-Plan (Plan-and-Solve).}
We append the instruction: \texttt{"Let's first plan how to solve this problem, then answer step by step."}

\subsection{\HierVA~Prompt Excerpts}
\label{sec:hva_prompt_excerpts}

We present shortened, paper-friendly excerpts of the manager prompts used by \HierVA{}. Full verbatim templates (including low-level formatting constraints) are omitted for readability.

\paragraph{Listing 1: Planning (manager).}
The manager requests a two-stage plan that decomposes the question into atomic, image-scoped sub-tasks, and optionally refines the plan with planning skills.

\begin{lstlisting}[caption={Manager planning prompt excerpt},label={lst:hva-plan}]
For the question '{question}', we need to develop a plan.
We can decompose the image and the question into sub-images + sub-tasks pairs.
Since the question can be complex, we need to plan twice.
The sub-image-task pair should be atomic.
Only plan now.
\end{lstlisting}

\paragraph{Listing 2: Task creation (manager).}
The manager specifies the next worker task by selecting the image input, writing a focused instruction, and declaring the expected output type. A review pass enforces a structured schema.

\begin{lstlisting}[caption={Task creation prompt excerpts},label={lst:hva-create}]
What is the next task? Provide for the sub-agent: the image to use,
the instruction, and the expected output type.

After reviewing, output in the following format:
- Sub-agent ID: [ID]
- Image: [image index]
- Instruction: [instruction]
- Known Information: [necessary information to finish the task]
- Tools: [image_zoom_in_tool, code_interpreter, or None]
- Skills: [choose from the skill allowlist or None]
- Expected Output: [image or text]
\end{lstlisting}

\paragraph{Listing 3: Dispatch + termination.}
Worker dispatch carries forward only distilled known information. Termination is decided by a yes/no check; final answers are standardized with a boxed wrapper.

\begin{lstlisting}[caption={Dispatch and termination prompt excerpts},label={lst:hva-dispatch}]
We know the following information: {known_information}.

Now let's assign the next task to the sub-agent. First check if there is
next task. Answer only "Yes" or "No".
Put your final answer in \boxed{}.
\end{lstlisting}

\paragraph{Listing 4: Tool schemas (injected in system prompt).}
The tool definitions are injected into the system prompt by the tool framework when a tool is enabled. Below we show the registered schema for the two tools used by \HierVA{}.

\begin{lstlisting}[caption={Tool prompt excerpts injected in the system prompt},label={lst:hva-tools}]
Tool: image_zoom_in_tool
Description: Zoom in on a specific region of an image by cropping it based on
             a bounding box (bbox) and an optional object label.
Parameters:
  - bbox_2d: array[4] of numbers, [x1, y1, x2, y2] with top-left and bottom-right
  - label: string, name or label of the object in the specified bbox
  - img_idx: number, index of the input image (starting from 0)

Tool: code_interpreter
Description: Python code sandbox, which can be used to execute Python code.
Parameters:
  - code: string, the python code
\end{lstlisting}

\section{Implementation Details}
\label{sec:impl_details}
This appendix provides additional engineering details needed to reproduce our system.
We specify the task schema used to communicate between manager and workers, our image indexing convention for chaining crops across steps, and the output contracts enforced for text and image tasks.

\subsection{Task Schema}
Each subtask at step $t$ is represented as
\[
\tau_t = (k_t,\ i_t,\ u_t,\ m_t,\ \mathcal{T}_t,\ \mathcal{S}_t,\ y_t),
\]
where $k_t$ is the worker index, $i_t$ selects the input image (original or a crop),
$u_t$ is the natural-language instruction, $m_t$ is a compact memory summary
(known information), $\mathcal{T}_t$ is the set of enabled tools, $\mathcal{S}_t$ is the set of injected skills,
and $y_t \in \{\textsc{Text},\textsc{Image}\}$ is the expected output type.

Manager-generated task specifications are validated against:
(i) required fields (non-empty instruction, declared expected output),
(ii) tool and skill allowlists, and
(iii) image reference validity.
If validation fails, we append a structured error message and re-prompt until a valid task is produced.

\subsection{Image Indexing Convention}
To enable multi-step visual reasoning without accumulating redundant images:
\begin{itemize}[noitemsep,topsep=0pt,leftmargin=1.25em]
    \item The original chart is referenced as image index $0$.
    \item When a worker with id $k$ produces a crop, that crop is stored as the \emph{latest image of worker $k$} and is referenced by setting \texttt{image\_index}$=k$ in later tasks.
\end{itemize}
This mechanism allows chaining zooms and reusing regions across subtasks.

\subsection{Output Contracts}
Workers produce outputs matching the expected type. For \texttt{text} tasks, we enforce a contract: the worker's final answer must appear in the last sentence, which is extracted and stored. For \texttt{image} tasks, the output is a crop handle. Skills may optionally declare lightweight \emph{hooks} for post-processing (e.g., coordinate normalization).

\end{document}